\documentclass{article} 
\usepackage[final]{colm2026_conference}

\usepackage{microtype}
\usepackage{hyperref}
\usepackage{url}
\usepackage{amsmath}
\usepackage{graphicx}
\usepackage{makecell}
\usepackage{booktabs}

\usepackage{lineno}

\definecolor{darkblue}{rgb}{0, 0, 0.5}
\hypersetup{colorlinks=true, citecolor=darkblue, linkcolor=darkblue, urlcolor=darkblue}

\definecolor{kh}{HTML}{168aff}

\title{Is CLIP Cross-Eyed? Revealing and Mitigating Center Bias in the CLIP Family}

\author{Oscar Chew$^1$, Hsiao-Ying Huang$^2$, Kunal Jain$^1$, Tai-I Chen$^{2,3}$, Khoa D. Doan$^4$, \\\bf Kuan-Hao Huang$^1$\\
$^1$Texas A\&M University, $^2$National Taiwan University, $^3$ASUS, $^4$VinUniversity\\
\texttt{oscarchew@tamu.edu}
}

\begin{document}

\ifcolmsubmission
\linenumbers
\fi

\maketitle

\begin{abstract}
Recent research has shown that contrastive vision-language models such as CLIP often lack a fine-grained understanding of visual content. While a growing body of work has sought to address this limitation, we identify a distinct failure mode in the CLIP family, which we term \textit{center bias}, that persists even in recent model variants. Specifically, CLIP tends to disproportionately focus on the central region of an image, overlooking important objects located near the boundaries. This limitation is fundamental as failure to recognize relevant objects makes it difficult to perform any sophisticated tasks that depend on those objects.  To understand the underlying causes of the limitation, we conduct analyses from both representation and attention perspectives. Using interpretability methods, i.e., embedding decomposition and attention map analysis, we find that relevant concepts especially those associated with off-center objects vanish from the model's embedding in the final representation due to information loss during the aggregation of visual embeddings, particularly the reliance on pooling mechanisms. Finally, we show that this bias can be alleviated with training-free strategies such as visual prompting and attention redistribution by redirecting models' attention to off-center regions. Our code is available at \url{https://github.com/lab-flair/vlm-center-bias}.
\end{abstract}

\section{Introduction}
Contrastive vision–language models (VLMs) such as CLIP \citep{radford2021learning} have become a foundational component in multimodal retrieval and generative systems. Despite their wide applications, recent studies have shown that CLIP and its variants often lack a fine-grained understanding of visual content. They often rely on a coarse understanding or spurious cues, failing to capture detailed object attributes, and exhibit ``bag-of-words'' behavior, struggling to accurately bind attributes to their corresponding objects \citep{yuksekgonul2023when, hsieh2023sugarcrepe,dumpala2024sugarcrepe++,tong2024eyes}.

\begin{figure}
    \centering
    \includegraphics[width=1\linewidth]{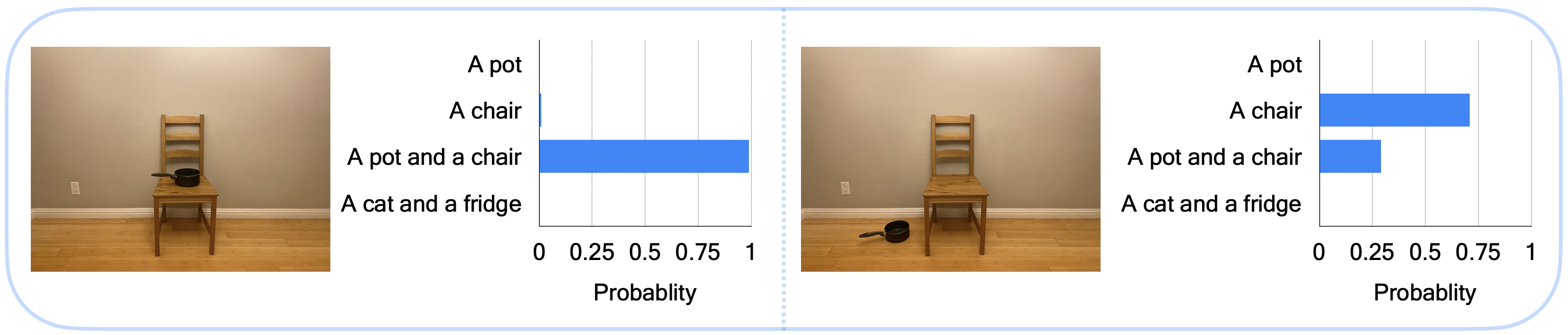}
    \caption{A representative example from \textbf{What'sUp}. When both objects are centrally aligned (left), CLIP correctly assigns high probability to “a pot and a chair”. However, when the pot is moved off-center (right), CLIP disproportionately focuses on the central chair, leading to a sharp drop in probability for the correct answer and a preference for “a chair.” 
    }
    \label{fig:failed-example}
\end{figure}

Beyond the inability to correctly associate present attributes, we observe a more critical failure mode: concepts can completely vanish from the model's embedding depending on where they appear in the image. 
Drawing inspiration from human vision research \citep{tseng2009quantifying,bindemann2010scene,borji2011quantifying}, which identifies a strong \emph{center bias} in gaze behavior driven by photographer bias and human viewing strategies, we examine whether analogous biases emerge in the CLIP family. As illustrated in Figure~\ref{fig:failed-example}, when an object is placed away from the center, the model may fail to recognize it altogether even when it is clearly visible. In this example, although both a chair (center) and a pot (off-center) are present, the model assigns high confidence to “a chair” while failing to capture the off-center object. Motivated by this observation, we reveal that CLIP tends to disproportionately focus on the central region of an image, systematically overlooking important objects located near the boundaries.

To systematically study this phenomenon, we evaluate model performance on both a repurposed real-world spatial relation dataset and a family of controlled synthetic datasets, demonstrating a consistent performance degradation on off-center objects across diverse model variants, including the latest advancements \citep{monsefi2025detailclip,zhao2025superclip}. Furthermore, we analyze the underlying causes of this bias from both representation and attention perspectives. Through embedding decomposition and attention map analysis, we find that relevant concepts associated with off-center objects are present in representations of other visual tokens but are not effectively captured in the final embedding. This indicates that the bias stems primarily from information loss during the aggregation into the \texttt{[CLS]} token, rather than an inherent lack of available visual information. 

Finally, having identified the mechanism behind this information loss, we demonstrate that this bias can be alleviated using training-free strategies. By employing visual prompting and attention redistribution, we can redirect the models' attention to off-center regions and recover overlooked concepts without retraining or modifying model parameters.

We summarize our contributions as follows: \begin{itemize}
\item We identify and quantify center bias in the CLIP family, showing consistent performance degradation on off-center objects across diverse datasets and model variants.
\item We provide analyses from both representation and attention perspectives, revealing the underlying mechanisms by which CLIP exhibits center bias.
\item We demonstrate that simple strategies, such as visual prompting and attention redistribution, can effectively alleviate center bias.
\end{itemize}

\section{Related Works}
\subsection{Contrastive Vision-Language Models}
Contrastive VLMs have emerged as a fundamental paradigm for aligning visual and textual representations. CLIP \citep{radford2021learning} learns a shared embedding space by contrasting matched image-text pairs against mismatched ones, enabling strong zero-shot transfer across a wide range of tasks. Subsequent works have improved this framework from different perspectives such as scaling to larger datasets and models \citep{cherti2023reproducible,chen2024vitamin}, integrating contrastive learning with additional pretraining tasks \citep{yu2022coca,eva02}, and adopting alternative training objectives \citep{Zhai_2023_ICCV}.
\subsection{Robust and Fine-grained Perception of VLMs}
Recent work has identified limitations in the fine-grained perception capabilities of the CLIP family. They show that CLIP often relies on coarse or spurious cues, failing to capture detailed object attributes and subtle visual distinctions \citep{yuksekgonul2023when,rahmanzadehgervi2024vision,adila2024zeroshot,wang2024sober,zhang-etal-2024-clip,tong2024eyes,koishigarina2026clip}. To address this, several methods have been introduced to improve fine-grained understanding. NegCLIP \citep{yuksekgonul2023when} incorporates hard negative examples to encourage better discrimination, while DetailCLIP \citep{monsefi2025detailclip} and SuperCLIP \citep{zhao2025superclip} augment contrastive learning with dense predictions or other auxiliary objectives to enhance sensitivity to fine-grained features.

In this work, we show that even recent CLIP variants continue to exhibit a systematic \emph{center bias}, indicating that improvements in fine-grained perception do not necessarily translate to better spatial coverage (Section~\ref{sec:revealing}).

\subsection{Center Bias in Human Vision}
Research in human vision has identified a strong center bias in gaze behavior, where observers tend to fixate near the center of visual scenes \citep{tseng2009quantifying,bindemann2010scene,borji2011quantifying}. This bias arises primarily from two factors: \emph{photographer bias}, where objects of interest are preferentially placed near the center of images, and \emph{viewing strategy}, where observers adopt a heuristic of initially attending to the center.

In this work, we draw inspiration from these findings and investigate whether similar biases emerge in contrastive vision-language models. In particular, we examine whether CLIP models exhibit a form of center bias analogous to human viewing strategy and dataset-driven biases, and whether this bias affects their ability to recognize objects outside the central region.


\section{Revealing Center Bias in CLIP}
In this section, we investigate the presence and extent of center bias in CLIP-based models. First, we repurpose the \textbf{What'sUp} dataset \citep{kamath-etal-2023-whats} and introduce a family of synthetic datasets, \textbf{GRID}, to isolate the effect of object position on model performance. We then establish a metric for center bias and benchmark a wide-range of CLIP variants. We hypothesize that despite advances in model scale, architecture, and training objectives, contrastive models will exhibit a significant and systematic performance degradation when recognizing objects placed outside the center of an image.

\subsection{Experiment Setup}
\paragraph{Datasets} First, we use \textbf{What'sUp} \citep{kamath-etal-2023-whats} subset A, where a single object is positioned relative to another object (e.g., on, left of, right of, or under a table, chair, or armchair). This dataset provides a real-world testbed for vision-language models. To connect this dataset with our study of positional bias, we partition the examples into two groups: \textbf{center} and \textbf{off-center}. We treat \textit{“on”} relations as the \textbf{center} set, since the primary object typically appears near the center of the image, while \textit{“left of,” “right of,”} and \textit{“under”} 
relations form the \textbf{off-center} set, where the primary object is more likely to be displaced from the center. We provide further justification for the validity of this setup in Appendix~\ref{sec:definition}.
Unlike the original What'sUp task, which requires predicting spatial relations, we instead consider a simpler recognition-based question: \textit{“What is in this image?”}. Surprisingly, we find that CLIP cannot reliably answer this easier question. The answer candidates include: (1) the primary object, (2) the supporting object (e.g., chair, table, or armchair), (3) both objects (the ground-truth answer), and (4) a distractor with a token length matched to the ground-truth answer. If CLIP struggles to answer such a simple recognition question, its predictions on more challenging spatial reasoning tasks may be unreliable.

\begin{figure}
    \centering
    \includegraphics[width=1\linewidth]{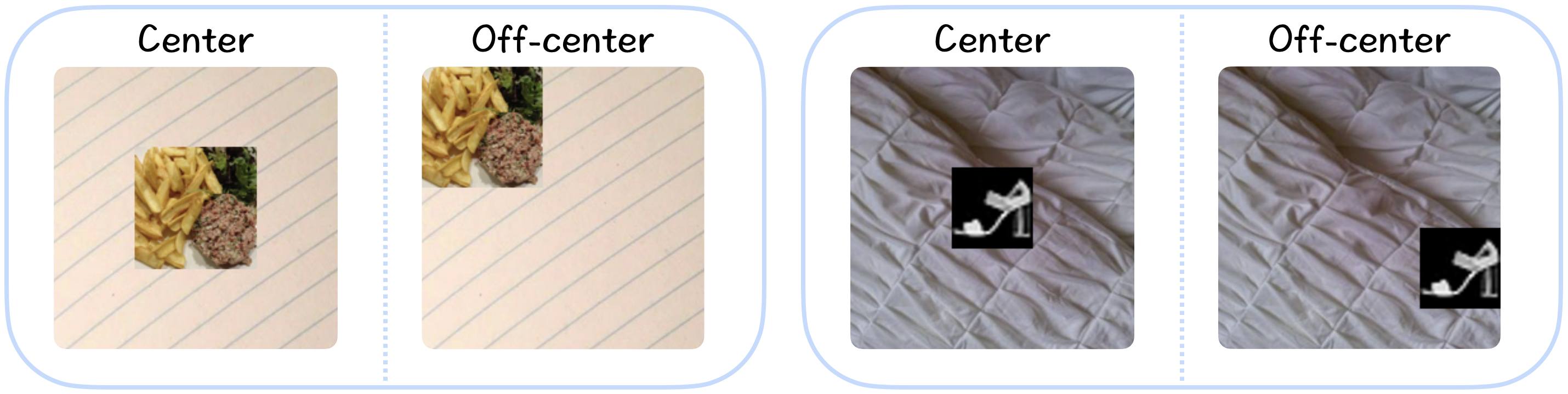}
    \caption{Examples of center and off-center configurations in the \textbf{GRID} dataset.}
    \label{fig:grid-example}
\end{figure}

Both object position and size affect whether a model can perceive objects in an image. To isolate the effect of position, we construct a family of synthetic datasets, \textbf{GRID}, while controlling for object size. These datasets are built from standard image classification benchmarks, including CIFAR-10 \citep{krizhevsky2009learning}, Fashion-MNIST \citep{xiao2017fashionmnist}, and Food-101 \citep{bossard2014food101}.
Each GRID sample consists of a larger canvas divided into a $k\times k$ grid, where a single object image, sized $s$=1/3/5 times the patch size, is placed in the grid. The backgrounds of the grids are chosen from a texture database \citep{cimpoi2014describing}. We define two subsets:
(1) \textbf{GRID-center}, where the object is always placed in the central cell, and
(2) \textbf{GRID-off-center}, where the object is randomly placed in the outer ring of the grid. This design allows us to isolate the effect of object position while keeping all other factors (e.g., object identity, scale, and background) consistent. The examples and construction details of \textbf{GRID} are provided in Figure~\ref{fig:grid-example} and Appendix~\ref{sec:dataset-details}.

\paragraph{Image preprocessing} We follow the standard preprocessing steps used in OpenCLIP. Specifically, every image undergoes the same procedure: resizing the shortest side to the target resolution (e.g., 224 pixels), center-cropping the image to (224 $\times$ 224), converting it to a tensor, and normalizing the tensor. This preprocessing does not affect GRID because all \textbf{GRID} images are already (224 $\times$ 224) squares. We also manually inspected every image in \textbf{What'sUp} after preprocessing and confirmed that the objects in all images remain visible.

\paragraph{Metrics}
We evaluate model performance using classification accuracy. In particular, we measure accuracy separately on the \textbf{center} and \textbf{off-center} subsets. \textit{center bias} is quantified as the performance gap between them. Our primary goal is to assess whether model performance is sensitive to object position. An ideal model without center bias should achieve high accuracy on both subsets while exhibiting a small performance gap between center and off-center cases.
\paragraph{Models}
We study a diverse set of CLIP-based models spanning different architectures, training objectives, scales, and input resolutions. Our evaluation includes OpenAI CLIP \citep{radford2021learning}, OpenCLIP \citep{cherti2023reproducible}, CoCa \citep{yu2022coca}, SigLIP \citep{Zhai_2023_ICCV}, EVA-02 \citep{eva02}, ViTamin \citep{chen2024vitamin} and variants designed to improve robustness or fine-grained understanding, such as NegCLIP \citep{yuksekgonul2023when}, DetailCLIP \citep{monsefi2025detailclip}, and SuperCLIP \citep{zhao2025superclip}.
\label{sec:revealing}

\subsection{Results}
\label{sec:downstream}

\paragraph{What'sUp} Table~\ref{tab:whatsup_initial} shows the performance of various CLIP-based models on the center and off-center subsets of \textbf{What'sUp}. Across all models, we observe a consistent and significant drop in accuracy when objects are placed away from the center. This degradation is not limited to a specific architecture or training strategy. It persists across different model families as well as recent variants designed to improve robustness or fine-grained perception. Improvements in scale or resolution do not consistently mitigate this issue. Larger models generally improve overall accuracy, but the gap between center and off-center performance remains substantial.

\begin{figure}
    \centering
    \includegraphics[width=0.48\linewidth]{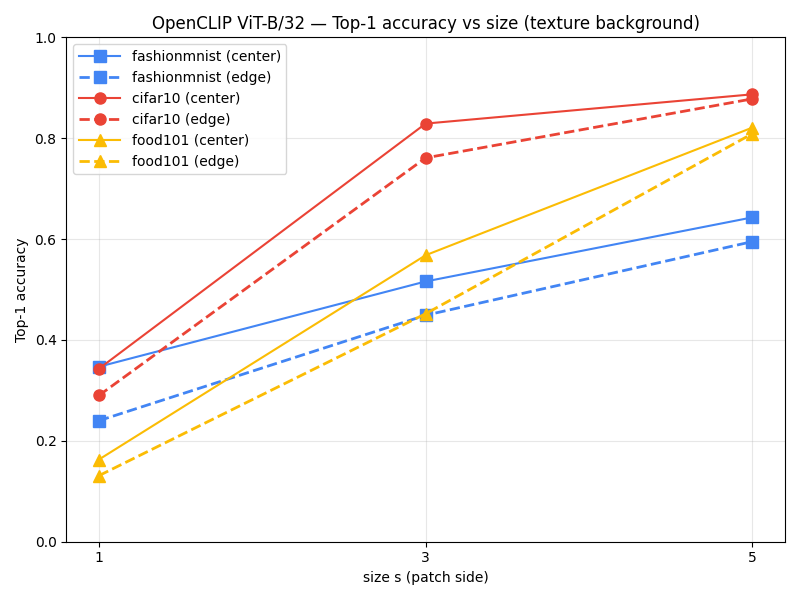}
    \includegraphics[width=0.48\linewidth]{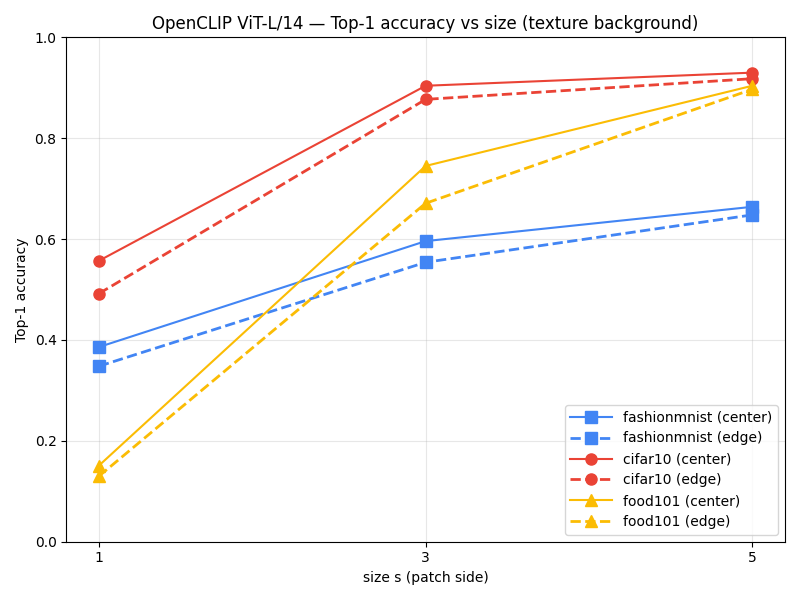}
    \caption{Mean top-1 accuracy vs. object size on a $7\times7$ \textbf{GRID}. Left: OpenCLIP ViT-B/32. Right: OpenCLIP ViT-L/14. Despite improvements with increasing object size, a persistent center–off-center performance gap remains.}
    \label{fig:gap-curve}
\end{figure}

\paragraph{GRID}  While \textbf{What'sUp} provides a realistic testbed, it also contains natural variations in object size and context. To isolate the effect of object location, we create the \textbf{GRID} datasets, where we fix the object size and vary its spatial position within a grid. An ideal object recognition model should be invariant to object location. However, as shown in Figure~\ref{fig:gap-curve}, even when object size is fixed, models consistently perform worse when the object is placed off-center.


\begin{table*}[h]
    \centering
    \begin{tabular}{lccc}
    \toprule
    Model & center ($\uparrow$) & off-center  ($\uparrow$)& center bias ($\downarrow$) \\
    \midrule
      OpenAI CLIP ViT-B/32 & 62.9 & 31.9 & 31.0 \\
      OpenAI CLIP ViT-L/14 & 94.2 & \textbf{76.7} & \textbf{17.5} \\
      OpenCLIP ViT-B/32 & 65.7 & 10.5 & 55.2 \\
      OpenCLIP ViT-L/14 & 87.6 & 46.0 & 41.6\\
      RoBERTaCLIP ViT-B/32 & 64.8 & 15.7 & 49.1\\
      CoCa ViT-B/32 & 39.0 & 6.70 & 32.3\\
      EVA02 ViT-B/16 & 56.2 & 13.7 & 42.5\\
      EVA02 ViT-L/14 & 88.6 & 35.1 & 53.5\\
      EVA02 ViT-L/14 (336) & 89.5 & 33.5 & 56.0\\
      SigLIP (512) & \textbf{97.1} & 69.6 & 27.5\\
      ViTamin-B & 66.7 & 34.5 & 32.2 \\
      ViTamin-L & 92.3 & 57.8 & 34.5 \\
      ViTamin-L (336) & \textbf{97.1} & 67.1 & 30.0 \\
      NegCLIP ViT-B-32 & 50.4 & 22.6 & 27.8\\
      DetailCLIP & 28.6 & 6.40 & 22.2 \\
      SuperCLIP ViT-B/16 & 50.5 & 22.4 & 28.1 \\
      SuperCLIP ViT-L/16 & 60.9 & 11.8 & 49.1 \\
    \bottomrule
    \end{tabular}
    \caption{Performance of CLIP-based models on the center and off-center subsets of \textbf{What'sUp}. Despite strong performance on centrally located objects, all models show significantly lower accuracy on off-center examples, indicating a systematic center bias.}
    \label{tab:whatsup_initial}
\end{table*}

These results collectively suggest that contrastive VLMs exhibit a systematic center bias, favoring centrally located content over equally relevant off-center information.

\section{Why Does CLIP Exhibit Center Bias?}
\label{sec:why}

Having established the presence of center bias across datasets and model variants in Section~\ref{sec:downstream}, we next seek to understand its underlying causes. While center bias is widespread, the exact mechanism may vary across architectures. In this work, we focus on class-token (\texttt{[CLS]})-based CLIP variants, one of the most important and popular designs (e.g., ViT, OpenCLIP models), where the aggregation process is particularly transparent to analyze.

To this end, we analyze CLIP from two complementary perspectives that capture both \emph{what} information is encoded and \emph{how} it is derived. First, we perform \textbf{embedding decomposition} to examine how different regions of an image contribute to the final representation and whether central regions disproportionately dominate the embedding. Second, we conduct \textbf{attention map analysis} to investigate whether the model inherently allocates more attention to central regions, potentially leading to the observed bias. 

\begin{figure}
    \centering
    \includegraphics[width=1\linewidth]{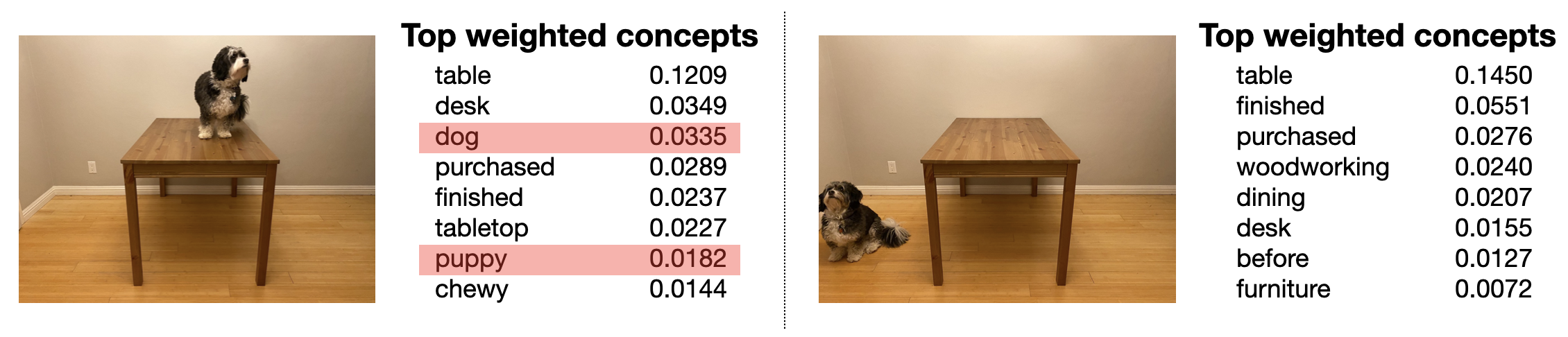}
    \caption{Comparison of top-weighted concepts from CLIP when the dog is placed at the center (left) versus off-center (right). When centered, CLIP assigns meaningful weights to ``dog''-related concepts, whereas these concepts disappear when the dog is placed off-center}
    \label{fig:concepts-missing}
\end{figure}
\subsection{Embedding Decomposition} As CLIP relies on a text-aligned vision encoder, we design the following experiment to examine whether its vision embeddings are sufficient for fine-grained understanding. We adopt SpLiCE \citep{bhalla2024interpreting}, an interpretable method that can decompose and explain the information captured by CLIP representations, and apply it to \textbf{What'sUp} examples.

Specifically, given an image $I$ and its CLIP vision embedding $\mathbf{x}$, we consider a set of concept keywords
$\{t_1, t_2, \cdots, t_n\}$ that may be present in the image. Let $\{\mathbf{c}_1, \mathbf{c}_2,\cdots, \mathbf{c}_n\}$ denote the corresponding CLIP text embeddings for these concept keywords and C denote a matrix with its $i$-th column being $\mathbf{c}_i$. SpLiCE then finds a set
of linear weights $\mathbf{w}$ over the concepts by solving the following optimization problem:

$$\min_{\mathbf{w}\in \mathcal{R}_+^n}||\mathbf{Cw}-\mathbf{x}||_2^2+2\lambda||\mathbf{{w}}||_1$$

Figure~\ref{fig:concepts-missing} illustrates an example where a table and a dog appear in both images. 
However, when the dog is moved from the center of the image (left figure) to the edge (right figure), the concept of ``dog'' is no longer among the highly weighted concepts. This suggests
that CLIP-based embeddings tend to encode a coarse-grained view of the image, emphasizing larger and centrally
located objects. Meanwhile, \textbf{the concepts of under-representing smaller or off-center elements vanish completely from the model's embedding}. More qualitative results for concept vanishing in other objects and positions pairs are provided in Appendix~\ref{sec:more-missing}.

\subsection{Attention Map Analysis}
A common approach to extract visual representations from ViTs, including many implementations of CLIP, is to use the representation of the \texttt{[CLS]} class token. This token is designed to aggregate information from all image patches through self-attention, and its final-layer embedding is typically used as the image representation. 

\begin{figure}
    \centering
    \includegraphics[width=1\linewidth]{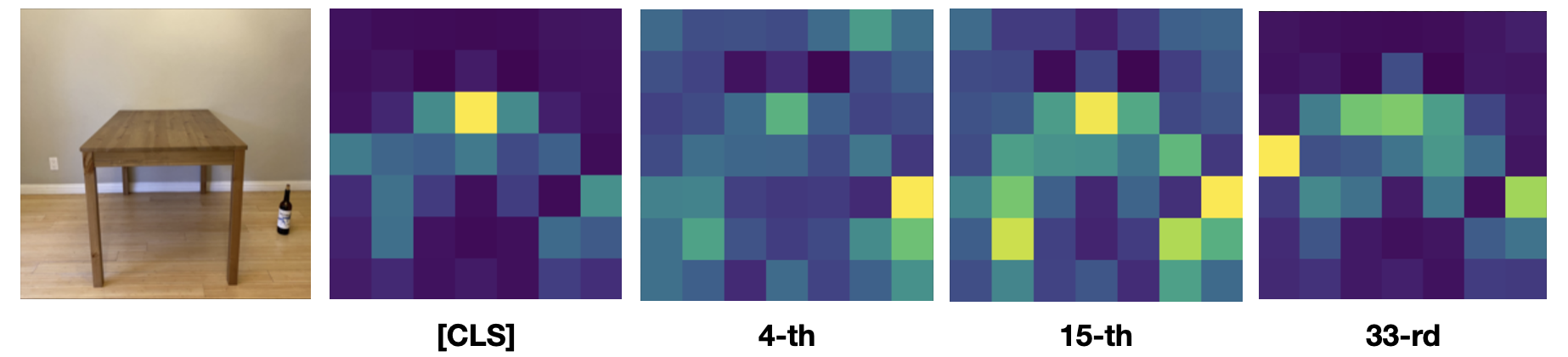}
    \caption{Attention maps for the \texttt{[CLS]} token and other visual tokens in the final layer. Brighter pixels indicate higher attention. While the \texttt{[CLS]} token focuses mainly on the table, patch tokens capture information from other important objects (the bottle and the socket). This suggests that center bias arises not from a lack of available information, but from information loss during pooling.}
    \label{fig:attention}
\end{figure}

However, in Figure~\ref{fig:attention}, we find that the attention used to compute the \texttt{[CLS]} embedding is excessively concentrated on the central region. While this behavior may suffice for the ordinary caption-matching task, it fails to capture fine-grained details, particularly for objects located near the boundaries. In contrast, other visual tokens still attend to these off-center objects. This suggests that \textbf{the relevant information is present in the intermediate representations but is not effectively preserved when aggregated into the \texttt{[CLS]} token.}

\begin{figure}
    \centering
    \includegraphics[width=1\linewidth]{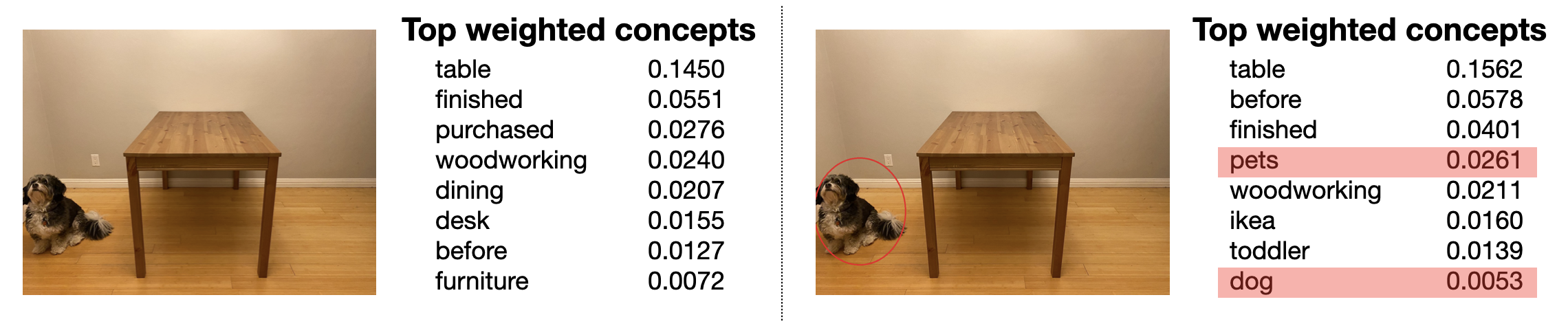}
    \caption{Effect of visual prompting on an off-center object. Without prompting (left), the model fails to assign meaningful ``dog''-related concepts. By adding a visual prompt (red circle) around the object (right), relevant concepts such as pets and dog reappear, indicating improved attention to the off-center region.}
    \label{fig:visprompts-reappear}
\end{figure}
\section{Test-time Mitigation of Center Bias}
As we have identified the heart of the problem is the information loss during pooling, we show two simple strategies namely visual prompting and attention redistribution to mitigate center bias for existing contrastive VLMs in this section. Visual prompting requires an external model but is general and applicable to all CLIP variants. Attention redistribution is empirically more consistent, although it is restricted to CLS-based CLIP models.

\label{sec:mitigation}
\subsection{Visual Prompting}
\label{sec:vp}
Prior work has shown that VLMs can be influenced by simple visual cues. \citet{Shtedritski_2023_ICCV} show that model attention can be redirected by drawing a red circle around objects of interest, effectively acting as a visual prompt. Motivated by this observation, we adopt a similar strategy to guide model attention. We use GroundingDINO \citep{liu2024grounding} to automatically detect any objects in the image, and then overlay red bounding boxes around the detected regions. These visual prompts encourage the model to attend to potentially overlooked regions, especially when objects are placed off-center. As shown in Figure~\ref{fig:visprompts-reappear}, adding a simple red circle around the object causes ``dog''-related concepts to reappear in the SpLiCE decomposition, even when the object is placed away from the center.

\begin{table*}[h]
    \centering
    \begin{tabular}{lcccc}
    \toprule
    Model & center ($\uparrow$) & off-center  ($\uparrow$)& \makecell[c]{center \\ bias ($\downarrow$)} & \makecell[c]{improv. \\ off-center ($\uparrow$)} \\
    \midrule
    \textbf{Visual Prompting} \\
    \midrule
      OpenAI CLIP ViT-B/32 & 54.2 & 21.8 & 32.4 & -10.1\\
      OpenAI CLIP ViT-L/14 & 87.6 & 76.7 & \textbf{10.9} & 0.00\\
      OpenCLIP ViT-B/32 & 59.0 & 23.3 & 35.7 & +12.8\\
      OpenCLIP ViT-L/14 & 81.0 & 56.5 & 24.5 & +10.5 \\
      RoBERTaCLIP ViT-B/32 & 52.3 & 21.7 & 30.6 & +6.00\\
      CoCa ViT-B/32 & 23.8 & 5.10 & 18.7 & -1.60\\
      EVA02 ViT-B/16 & 65.7 & 33.8 & 31.9 & +20.1  \\
      EVA02 ViT-L/14 & 86.6 & 69.0 & 17.6 & +33.9\\
      EVA02 ViT-L/14 (336) & 91.4 & 73.5 & 17.9 & +40.0\\
      SigLIP (512) & 88.5 & 70.9 & 17.6 & +1.30  \\
      ViTamin-B & 56.1 & 38.3 & 17.8 & +3.80  \\
      ViTamin-L & 79.0 & 56.8 & 22.2 & -1.00 \\
      ViTamin-L (336) & 88.5 & 70.9 & 17.6 & +3.80 \\
      NegCLIP ViT-B-32 & 37.1 & 14.1 & 23.0 & -8.50 \\
      DetailCLIP & 20.0 & 10.9 & 9.10 & +4.50 \\
      SuperCLIP ViT-B/16 & 52.4 & 30.0 & 22.4 & +7.60 \\
      SuperCLIP ViT-L/16 & 50.5 & 20.1 & 30.4 & +8.30\\
      \midrule
      \textit{mean performance} & 63.2 & 40.8 & 22.4 & +7.71 \\
    \toprule
    \textbf{Attention Redistribution} \\
    \midrule
        OpenAI CLIP ViT-B/32 & 76.2 & 49.2 & 27.0 & +17.3 \\
      OpenAI CLIP ViT-L/14 & \textbf{94.3} & \textbf{77.3} & 17.0 & +0.60\\
      OpenCLIP ViT-B/32 &  68.6 & 25.2 & 43.4 & +14.7\\
      OpenCLIP ViT-L/14 &  89.5 & 48.9 & 40.6 & +2.90 \\
      \midrule
      \textit{mean performance} & 82.2 & 50.2 & 32.0 & +8.90 \\
    \bottomrule
    \end{tabular}
    \caption{Performance comparison of visual prompting (VP) and attention redistribution (AR) across various CLIP variants on \textbf{What'sUp}. VP generally improves off-center performance but exhibits inconsistent effects across models. In contrast, AR improves off-center performance and overall accuracy more consistently.}
    \label{tab:mitigation-whatsup}
\end{table*}

\subsection{Attention Redistribution}
Given our observation in Section~\ref{sec:why}, a natural way to mitigate center bias is to intervene on the final attention used to form the \texttt{[CLS]} representation. While the \texttt{[CLS]} token assigns disproportionately high attention on the central object, other visual tokens often attend to a broader set of relevant regions, including off-center objects. 
Motivated by this, we redistribute the final-layer attention of the \texttt{[CLS]} token by suppressing its self-attention and renormalizing the remaining mass over patch tokens. Let $A \in \mathbb{R}^{(N+1)\times(N+1)}$ denote the final-layer attention matrix, where index $0$ corresponds to the \texttt{[CLS]} token and $N$ is the number of patches. We set

\[
\tilde{A}_{0,0}=0,
\]
and renormalize the rest of the \texttt{[CLS]} row:
\[
\tilde{A}_{0,j} = \frac{A_{0,j}}{\sum_{k=1}^{N} A_{0,k}}, \quad j=1,\dots,N.
\]
This modification preserves the relative importance among visual tokens while forcing \texttt{[CLS]} to rely more on patch-level evidence, which in turn reduces the dominance of the center region. We apply attention redistribution to CLS-based CLIP implementations available in OpenCLIP.

An alternative idea is to directly aggregate representations from other patch tokens, for example by taking the mean representation. However, this approach does not work, as the projection layer in CLIP is trained specifically on the \texttt{[CLS]} representation and does not generalize well to other aggregation schemes.


\subsection{Results}
Tables~\ref{tab:mitigation-whatsup} and~\ref{tab:mitigation-grid} present the results of visual prompting and attention redistribution on \textbf{What'sUp} and \textbf{GRID}. Both methods yield net improvements, achieving average gains of 7.71 and 8.90 percentage points in off-center performance, respectively, compared to their vanilla counterparts in the real-world \textbf{What'sUp} dataset.

However, the two approaches exhibit different trade-offs. Visual prompting is applicable to all CLIP variants, but on both datasets, despite its ability to reduce center bias, visual prompting sometimes degrades overall accuracy. In contrast, attention redistribution consistently improves off-center performance, while maintaining or improving overall accuracy, but only suitable for class-token based CLIP variants.


These results support our hypothesis that center bias stems from the feature aggregation mechanism, and that redistributing attention toward patch tokens better highlights off-center information. Nevertheless, as discussed in Section~\ref{sec:vp}, visual prompting remains a flexible, model-agnostic approach that can be easily applied to any existing architectures.

\begin{table*}[h]
    \centering
    \begin{tabular}{lccc}
    \toprule
    Model & center ($\uparrow$) & off-center  ($\uparrow$)& \makecell[c]{center \\ bias ($\downarrow$)} \\
    \midrule
      OpenAI CLIP ViT-B/32 & 19.1 & \textbf{15.7} & \textbf{3.40}\\
      $\quad+$ VP & 17.7 & 13.2 & 4.50\\
      $\quad+$ AR & \textbf{19.2} & \textbf{15.7} & 3.50\\
      \midrule
      OpenAI CLIP ViT-L/14 & 30.5 & 26.0 & \textbf{4.50}\\
        $\quad+$ VP & 30.4 & 25.2 & 5.2 \\
      $\quad+$ AR & \textbf{31.2} & \textbf{26.7} & \textbf{4.50}\\
      \midrule
      OpenCLIP ViT-B/32 & 17.2 & 15.1 & \textbf{2.07} \\
      $\quad+$ VP & 15.7 & 12.7 & 3.03\\
      $\quad+$ AR & \textbf{18.6}& \textbf{16.5} & \textbf{2.07}\\
      \midrule
      OpenCLIP ViT-L/14 & 35.2 & 31.4 & 3.83\\
      $\quad+$ VP & 34.4 & 27.9 & 6.50\\
      $\quad+$ AR & \textbf{35.3} & \textbf{31.7} & \textbf{3.53} \\
      \midrule
    \textit{mean model} & 25.5 & 22.1 & 3.45 \\
    $\quad+$VP & 24.6 & 19.8 & 4.80 \\
    $\quad+$AR & \textbf{26.0} & \textbf{22.7} & \textbf{3.43} \\
    \bottomrule
    \end{tabular}
    \caption{Mean performance across CIFAR10, FashionMNIST, and Food101 comparing visual prompting (VP) and attention redistribution (AR) on the synthetic $7\times7$ \textbf{GRID} with $s=1$. AR improves both off-center performance and overall accuracy.}
    \label{tab:mitigation-grid}
\end{table*}


\section{Conclusion}
In this paper, we identify a systematic center bias in the CLIP family, where models excessively focus on centrally located content while overlooking off-center objects. Through experiments across multiple datasets and model variants, we show that this bias leads to consistent performance degradation.

We further analyze this phenomenon for the representative class token-based CLIP from both representation and attention perspectives, revealing that relevant information from off-center regions is not effectively preserved due to information loss during pooling. To alleviate this issue, we propose simple test-time strategies, including visual prompting and attention redistribution, which encourage the model to attend to underrepresented regions. Experimental results demonstrate that these approaches can alleviate center bias without modifying model parameters. 

Overall, our findings highlight a potential root cause of perceptual blindness in VLMs and motivate a broader re-examination of representation to enable more reliable and spatially aware visual understanding.

\section{Limitations and Future Work}
In this work, we focus on revealing and understanding center bias in existing CLIP models and propose simple test-time intervention techniques to alleviate it. While our approaches are relatively simple, we believe the findings could inspire future work to explore more sophisticated test-time strategies for mitigating center bias.

Another promising next step is to prevent the emergence of such biases during the contrastive pretraining stage itself, for example by incorporating stronger augmentations or designing position-invariant objectives. 

Finally, our analysis reveals that center bias is widespread across CLIP variants and not unique to class token-based architectures. Understanding how center bias emerges in these alternative architectures is an important direction.

\section*{Acknowledgement}
This research is partially supported by the National Artificial Intelligence Research Resource Pilot (NAIRR) program (No. NAIRR260216) and Advanced Cyberinfrastructure Coordination Ecosystem: Services \& Support (ACCESS) program (No. CIS251227). Portions of this research were conducted with the advanced computing resources provided by Texas A\&M High Performance Research Computing.

\bibliography{colm2026_conference}

@InProceedings{radford2021learning,
  title = 	 {Learning Transferable Visual Models From Natural Language Supervision},
  author =       {Radford, Alec and Kim, Jong Wook and Hallacy, Chris and Ramesh, Aditya and Goh, Gabriel and Agarwal, Sandhini and Sastry, Girish and Askell, Amanda and Mishkin, Pamela and Clark, Jack and Krueger, Gretchen and Sutskever, Ilya},
  booktitle = 	 {Proceedings of the 38th International Conference on Machine Learning},
  pages = 	 {8748--8763},
  year = 	 {2021},
  editor = 	 {Meila, Marina and Zhang, Tong},
  volume = 	 {139},
  series = 	 {Proceedings of Machine Learning Research},
  month = 	 {18--24 Jul},
  publisher =    {PMLR},
  url = 	 {https://proceedings.mlr.press/v139/radford21a.html},
}

@InProceedings{Shtedritski_2023_ICCV,
    author    = {Shtedritski, Aleksandar and Rupprecht, Christian and Vedaldi, Andrea},
    title     = {What does CLIP know about a red circle? Visual prompt engineering for VLMs},
    booktitle = {Proceedings of the IEEE/CVF International Conference on Computer Vision (ICCV)},
    month     = {October},
    year      = {2023},
    pages     = {11987-11997}
}

@inproceedings{cherti2023reproducible,
  title={Reproducible scaling laws for contrastive language-image learning},
  author={Cherti, Mehdi and Beaumont, Romain and Wightman, Ross and Wortsman, Mitchell and Ilharco, Gabriel and Gordon, Cade and Schuhmann, Christoph and Schmidt, Ludwig and Jitsev, Jenia},
  booktitle={Proceedings of the IEEE/CVF Conference on Computer Vision and Pattern Recognition},
  pages={2818--2829},
  year={2023}
}

@inproceedings{bhalla2024interpreting,
 author = {Bhalla, Usha and Oesterling, Alex and Srinivas, Suraj and Calmon, Flavio P. and Lakkaraju, Himabindu},
 booktitle = {Advances in Neural Information Processing Systems},
 editor = {A. Globerson and L. Mackey and D. Belgrave and A. Fan and U. Paquet and J. Tomczak and C. Zhang},
 pages = {84298--84328},
 publisher = {Curran Associates, Inc.},
 title = {Interpreting CLIP with Sparse Linear Concept Embeddings (SpLiCE)},
 url = {https://proceedings.neurips.cc/paper_files/paper/2024/file/996bef37d8a638f37bdfcac2789e835d-Paper-Conference.pdf},
 volume = {37},
 year = {2024}
}

@inproceedings{kamath-etal-2023-whats,
    title = "What`s {\textquotedblleft}up{\textquotedblright} with vision-language models? Investigating their struggle with spatial reasoning",
    author = "Kamath, Amita  and
      Hessel, Jack  and
      Chang, Kai-Wei",
    editor = "Bouamor, Houda  and
      Pino, Juan  and
      Bali, Kalika",
    booktitle = "Proceedings of the 2023 Conference on Empirical Methods in Natural Language Processing",
    month = dec,
    year = "2023",
    address = "Singapore",
    publisher = "Association for Computational Linguistics",
    url = "https://aclanthology.org/2023.emnlp-main.568/",
    doi = "10.18653/v1/2023.emnlp-main.568",
    pages = "9161--9175"
}

@inproceedings{liu2024grounding,
  title={Grounding dino: Marrying dino with grounded pre-training for open-set object detection},
  author={Liu, Shilong and Zeng, Zhaoyang and Ren, Tianhe and Li, Feng and Zhang, Hao and Yang, Jie and Li, Chunyuan and Yang, Jianwei and Su, Hang and Zhu, Jun and others},
  booktitle={European Conference on Computer Vision},
  year={2024}
}

@article{eva02,
  title={Eva-02: A visual representation for neon genesis},
  author={Fang, Yuxin and Sun, Quan and Wang, Xinggang and Huang, Tiejun and Wang, Xinlong and Cao, Yue},
  journal={Image and Vision Computing},
  pages={105171},
  year={2024},
  publisher={Elsevier}
}

@InProceedings{Zhai_2023_ICCV,
    author    = {Zhai, Xiaohua and Mustafa, Basil and Kolesnikov, Alexander and Beyer, Lucas},
    title     = {Sigmoid Loss for Language Image Pre-Training},
    booktitle = {Proceedings of the IEEE/CVF International Conference on Computer Vision (ICCV)},
    month     = {October},
    year      = {2023},
    pages     = {11975-11986}
}

@inproceedings{chen2024vitamin,
  title={ViTamin: Designing Scalable Vision Models in the Vision-language Era},
  author={Chen, Jieneng and Yu, Qihang and Shen, Xiaohui and Yuille, Alan and Chen, Liang-Chieh},
  booktitle={Proceedings of the IEEE/CVF Conference on Computer Vision and Pattern Recognition},
  year={2024}
}

@inproceedings{
yuksekgonul2023when,
title={When and Why Vision-Language Models Behave like Bags-Of-Words, and What to Do About It?},
author={Mert Yuksekgonul and Federico Bianchi and Pratyusha Kalluri and Dan Jurafsky and James Zou},
booktitle={The Eleventh International Conference on Learning Representations },
year={2023},
url={https://openreview.net/forum?id=KRLUvxh8uaX}
}

@inproceedings{
monsefi2025detailclip,
title={Detail{CLIP}: Detail-Oriented {CLIP} for Fine-Grained Tasks},
author={Amin Karimi Monsefi and Kishore Prakash Sailaja and Ali Alilooee and Ser-Nam Lim and Rajiv Ramnath},
booktitle={Scaling Self-Improving Foundation Models without Human Supervision},
year={2025},
url={https://openreview.net/forum?id=80AuuQK4UQ}
}

@inproceedings{
zhao2025superclip,
title={Super{CLIP}: {CLIP} with Simple Classification Supervision},
author={Weiheng Zhao and Zilong Huang and Jiashi Feng and Xinggang Wang},
booktitle={The Thirty-ninth Annual Conference on Neural Information Processing Systems},
year={2025},
url={https://openreview.net/forum?id=EeIEvZlmVg}
}

@inproceedings{krizhevsky2009learning,
  title={Learning multiple layers of features from tiny images},
  author={Krizhevsky, Alex and Hinton, Geoffrey},
  booktitle={Technical Report, Computer Science Department, University of Toronto},
  year={2009},
}

@online{xiao2017fashionmnist,
  author       = {Han Xiao and Kashif Rasul and Roland Vollgraf},
  title        = {Fashion-MNIST: a Novel Image Dataset for Benchmarking Machine Learning Algorithms},
  date         = {2017-08-28},
  year         = {2017},
  eprintclass  = {cs.LG},
  eprinttype   = {arXiv},
  eprint       = {cs.LG/1708.07747},
}

@inproceedings{bossard2014food101,
  title = {Food-101 -- Mining Discriminative Components with Random Forests},
  author = {Bossard, Lukas and Guillaumin, Matthieu and Van Gool, Luc},
  booktitle = {European Conference on Computer Vision},
  year = {2014}
}

@article{
yu2022coca,
title={CoCa: Contrastive Captioners are Image-Text Foundation Models},
author={Jiahui Yu and Zirui Wang and Vijay Vasudevan and Legg Yeung and Mojtaba Seyedhosseini and Yonghui Wu},
journal={Transactions on Machine Learning Research},
issn={2835-8856},
year={2022},
url={https://openreview.net/forum?id=Ee277P3AYC},
note={}
}

@article{tseng2009quantifying,
  title={Quantifying center bias of observers in free viewing of dynamic natural scenes},
  author={Tseng, Po-He and Carmi, Ran and Cameron, Ian GM and Munoz, Douglas P and Itti, Laurent},
  journal={Journal of vision},
  volume={9},
  number={7},
  pages={4--4},
  year={2009},
  publisher={The Association for Research in Vision and Ophthalmology}
}

@article{borji2011quantifying,
  title={Quantifying the relative influence of photographer bias and viewing strategy on scene viewing},
  author={Borji, Ali and Sihite, Dicky N and Itti, Laurent},
  journal={Journal of Vision},
  volume={11},
  number={11},
  pages={166},
  year={2011},
  publisher={The Association for Research in Vision and Ophthalmology}
}

@article{bindemann2010scene,
  title={Scene and screen center bias early eye movements in scene viewing},
  author={Bindemann, Markus},
  journal={Vision research},
  volume={50},
  number={23},
  pages={2577--2587},
  year={2010},
  publisher={Elsevier}
}

@inproceedings{tong2024eyes,
  title={Eyes wide shut? exploring the visual shortcomings of multimodal llms},
  author={Tong, Shengbang and Liu, Zhuang and Zhai, Yuexiang and Ma, Yi and LeCun, Yann and Xie, Saining},
  booktitle={Proceedings of the IEEE/CVF conference on computer vision and pattern recognition},
  pages={9568--9578},
  year={2024}
}

@article{hsieh2023sugarcrepe,
  title={Sugarcrepe: Fixing hackable benchmarks for vision-language compositionality},
  author={Hsieh, Cheng-Yu and Zhang, Jieyu and Ma, Zixian and Kembhavi, Aniruddha and Krishna, Ranjay},
  journal={Advances in neural information processing systems},
  volume={36},
  pages={31096--31116},
  year={2023}
}

@article{dumpala2024sugarcrepe++,
  title={Sugarcrepe++ dataset: Vision-language model sensitivity to semantic and lexical alterations},
  author={Dumpala, Sri H and Jaiswal, Aman and Sastry, Chandramouli and Milios, Evangelos and Oore, Sageev and Sajjad, Hassan},
  journal={Advances in Neural Information Processing Systems},
  volume={37},
  pages={17972--18018},
  year={2024}
}

@InProceedings{cimpoi2014describing,
author = {Cimpoi, Mircea and Maji, Subhransu and Kokkinos, Iasonas and Mohamed, Sammy and Vedaldi, Andrea},
title = {Describing Textures in the Wild},
booktitle = {Proceedings of the IEEE Conference on Computer Vision and Pattern Recognition (CVPR)},
month = {June},
year = {2014}
}

@article{wang2024sober,
  title={A sober look at the robustness of clips to spurious features},
  author={Wang, Qizhou and Lin, Yong and Chen, Yongqiang and Schmidt, Ludwig and Han, Bo and Zhang, Tong},
  journal={Advances in Neural Information Processing Systems},
  volume={37},
  pages={122484--122523},
  year={2024}
}

@inproceedings{rahmanzadehgervi2024vision,
  title={Vision language models are blind},
  author={Rahmanzadehgervi, Pooyan and Bolton, Logan and Taesiri, Mohammad Reza and Nguyen, Anh Totti},
  booktitle={Proceedings of the Asian Conference on Computer Vision},
  pages={18--34},
  year={2024}
}

@inproceedings{
adila2024zeroshot,
title={Zero-Shot Robustification of Zero-Shot Models},
author={Dyah Adila and Changho Shin and Linrong Cai and Frederic Sala},
booktitle={The Twelfth International Conference on Learning Representations},
year={2024},
url={https://openreview.net/forum?id=fCeUoDr9Tq}
}

@inproceedings{zhang-etal-2024-clip,
    title = "Can {CLIP} Count Stars? An Empirical Study on Quantity Bias in {CLIP}",
    author = "Zhang, Zeliang  and
      Liu, Zhuo  and
      Feng, Mingqian  and
      Xu, Chenliang",
    editor = "Al-Onaizan, Yaser  and
      Bansal, Mohit  and
      Chen, Yun-Nung",
    booktitle = "Findings of the Association for Computational Linguistics: EMNLP 2024",
    month = nov,
    year = "2024",
    address = "Miami, Florida, USA",
    publisher = "Association for Computational Linguistics",
    url = "https://aclanthology.org/2024.findings-emnlp.59/",
    doi = "10.18653/v1/2024.findings-emnlp.59",
    pages = "1081--1086"
}

@inproceedings{
koishigarina2026clip,
title={{CLIP} Behaves like a Bag-of-Words Model Cross-modally but not Uni-modally},
author={Darina Koishigarina and Arnas Uselis and Seong Joon Oh},
booktitle={The Fourteenth International Conference on Learning Representations},
year={2026},
url={https://openreview.net/forum?id=DldwXCCP25}
}
\bibliographystyle{colm2026_conference}

\newpage
\appendix
\section{Dataset Details}
\label{sec:dataset-details}
\paragraph{What'sUp}
We use all 418 images from \textbf{What'sUp} Subset A, of which 105 are \textit{center} and 313 are \textit{off-center}.

\paragraph{GRID}
For each of the three datasets (FashionMNIST, CIFAR-10, and Food101), we randomly select 10 classes and sample 100 images per class from the selected classes to form the set of base images. For each base image, we generate two instances: one placed at the center and one placed at a random position along the outer ring. This results in 6,000 examples for each fixed object size $s$. We consider $s \in \{1, 3, 5\}$ to ensure that the center position is well-defined while preventing the object from exceeding the intended patch boundaries.

\section{On the definition of \textit{center} in What'sUp}
\label{sec:definition}
We justify our definitions of \textit{center} and \textit{off-center} by comparing the mean and median distances between the bounding-box centroids and the image center for each position, normalized by the image diagonal length. On average, the “on” position has a significantly smaller distance than the other positions, supporting the validity of our experimental setup.

\begin{table}[!ht]
    \centering
    \begin{tabular}{lcc}
    \toprule
        Position & Mean Distance & Median Distance \\ \midrule
        Left & 0.273 & 0.288 \\ 
        Right & 0.261 & 0.276 \\
        Under & 0.192 & 0.198 \\
        On & \textbf{0.064} & \textbf{0.060} \\
        \bottomrule
    \end{tabular}
\end{table}

We also tested the setting where "under" is removed from the off-center cases. The results show no meaningful change, suggesting that "under" behaves similarly to "left" and "right".

\begin{table}[!ht]
    \centering
    \begin{tabular}{lccc}
    \toprule
        Model & Center & Off-center & Center-bias \\ \midrule
        OpenAI CLIP ViT-B/32 & 62.9 & 32.9 & 30.00 \\ 
        OpenAI CLIP ViT-L/14 & 94.2 & 80.4 & 13.8 \\
        OpenCLIP ViT-B/32 & 65.7 & 8.0 & 57.7 \\
        OpenCLIP ViT-L/14 & 87.6 & 41.9 & 45.7 \\ \bottomrule
    \end{tabular}
\end{table}

\begin{figure}
    \centering
    \includegraphics[width=1\linewidth]{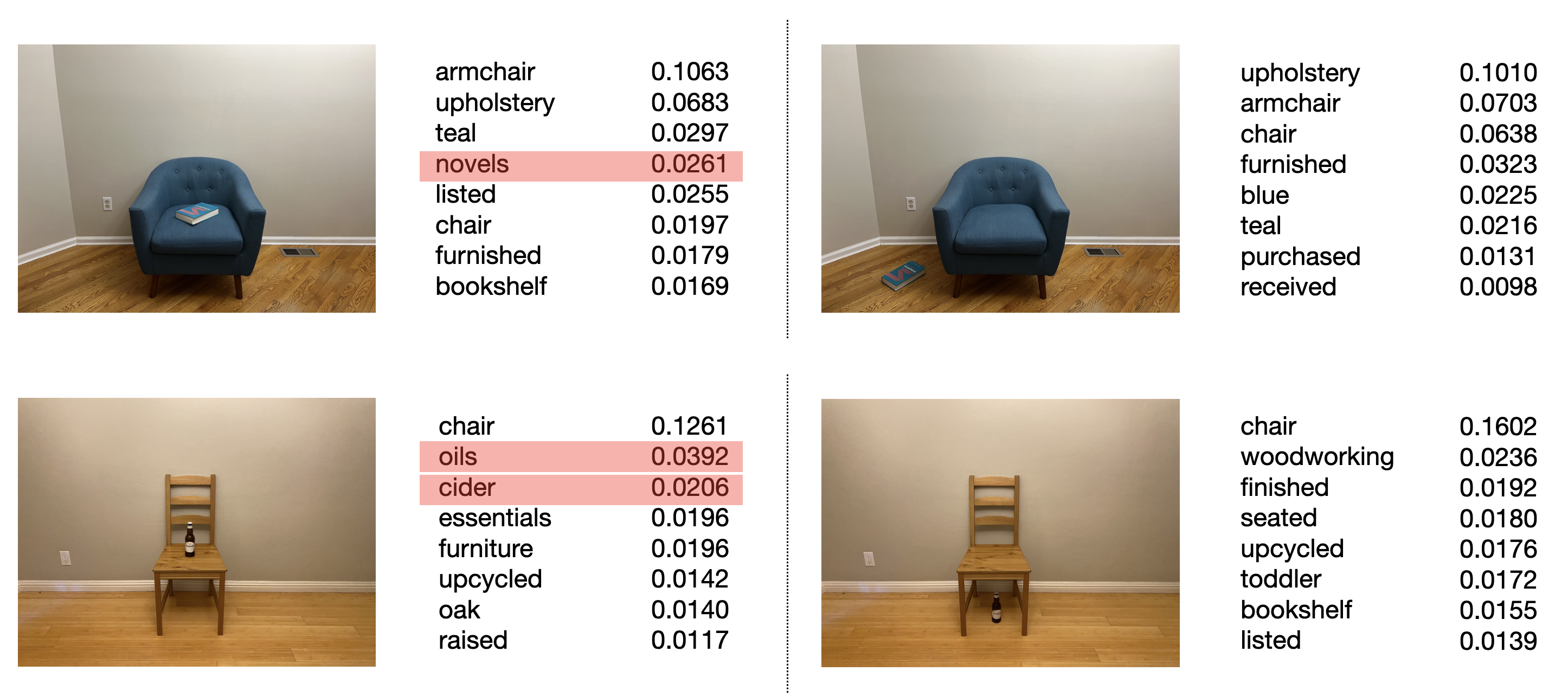}
    \caption{Additional examples of concept vanishing under different object positions.}
    \label{fig:missing2}
\end{figure}

\section{More Examples of Concept Vanishing}
\label{sec:more-missing}

We provide additional qualitative examples of 'concept vanishing' across varying spatial positions (center vs. left/under) in Figure~\ref{fig:missing2} for \textbf{What'sUp}. Consistent with our findings in the main text, the model’s representation is heavily biased toward the central object (e.g., armchair, chair), leaving off-center elements under-represented. 

\begin{figure}
    \centering
    \includegraphics[width=1\linewidth]{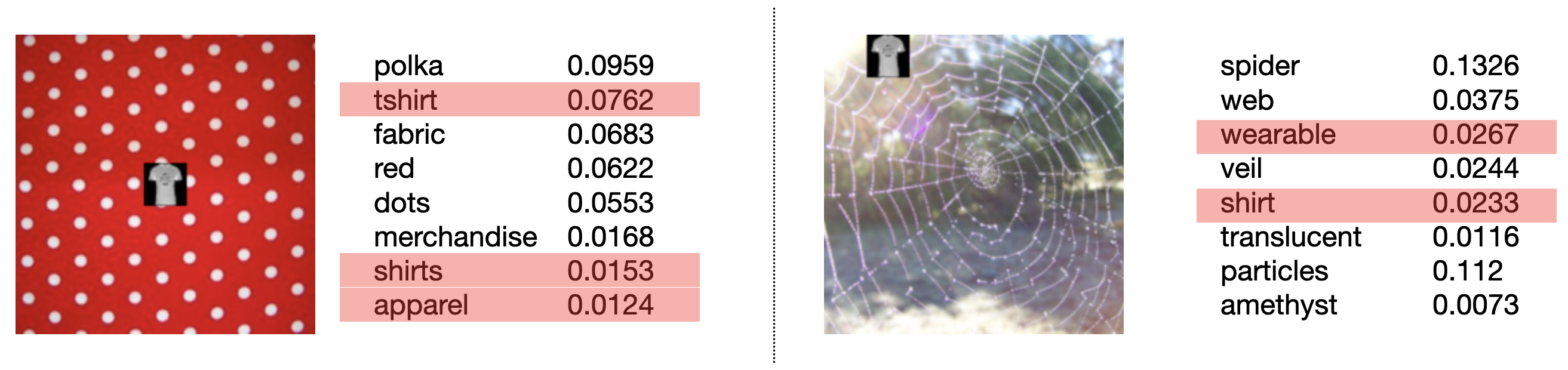}
    \caption{Additional examples of concept vanishing under different object positions.}
    \label{fig:missing3}
\end{figure}

The general trend remains the same on GRID. As the object moves from the center to off-center regions, the corresponding concept becomes less represented in the final embedding, and its coefficient and rank in the decomposition analysis decrease (Figure~\ref{fig:missing3}).

\end{document}